\documentclass{article}
\usepackage{spconf,amsmath,graphicx}

\usepackage[sort&compress]{natbib} % im text nach nummer sortierte literaturverweise
\bibpunct{[}{]}{,}{n}{,}{,}
\usepackage{url}
\usepackage[ngerman, american]{babel}
\newcommand{\eg}{\textit{e.g.}}
\newcommand{\ie}{\textit{i.e.}}

\usepackage{cleveref}
\crefname{appendix}{App.\negthinspace\,}{App.\negthinspace\,}
\crefname{chapter}{Chap.\negthinspace\,}{Chap.\negthinspace\,}
\crefname{equation}{Eq.\negthinspace\,}{Eq.\negthinspace\,}
\crefname{algorithm}{Alg.\negthinspace\,}{Alg.\negthinspace\,}
\crefname{section}{Sec.\negthinspace\,}{Sec.\negthinspace\,}
\crefname{subsection}{Sec.\negthinspace\,}{Sec.\negthinspace\,}
\crefname{subsubsection}{Sec.\negthinspace\,}{Sec.\negthinspace\,}
\crefname{figure}{Fig.\negthinspace\,}{Fig.\negthinspace\,}
\crefname{table}{Tab.\negthinspace\,}{Tab.\negthinspace\,}
\crefname{subfigure}{Fig.\negthinspace\,}{Fig.\negthinspace\,}
\crefname{subsubfigure}{Fig.\negthinspace\,}{Fig.\negthinspace\,}
\crefname{lstlisting}{Lst.\negthinspace\,}{Lst.\negthinspace\,}

\title{Generating Semi-Synthetic Validation Benchmarks for Embryomics}

\name{{\parbox[c]{\textwidth}{\centering Johannes Stegmaier$^{\star}$ \qquad Julian~Arz$^\ddagger$ \qquad Benjamin Schott$^{\star}$ \qquad Jens C. Otte$^\dagger$ \qquad Andrei Kobitski$^*$ \\  \qquad G.~Ulrich~Nienhaus$^{*,\circ}$ \qquad Uwe~Str\"ahle$^\dagger$ \qquad Peter~Sanders$^\ddagger$ \qquad Ralf~Mikut$^{\star}$\thanks{We are grateful for funding by the Helmholtz Association in the program BioInterfaces (BS, JO, US, RM) and Science and Technology of Nanosystems (AK, GUN), the Karlsruhe Institute of Technology (JA, PS) and the German Research Foundation DFG in the project MI1315/4-1 (JS).}}}}

\address{
$^{\star}$Institute for Applied Computer Science, Karlsruhe Institute of Technology, Karlsruhe, Germany\\
$^\ddagger$Institute of Theoretical Informatics, Karlsruhe Institute of Technology, Karlsruhe, Germany\\
$^\dagger$Institute of Toxicology and Genetics, Karlsruhe Institute of Technology, Karlsruhe, Germany\\ 
$^*$Institute of Applied Physics, Karlsruhe Institute of Technology, Karlsruhe, Germany\\
$^\circ$Department of Physics, University of Illinois at Urbana-Champaign, Urbana, Illinois 61801, USA}
\begin{document}
%\ninept
%
\maketitle
\begin{abstract}
Systematic validation is an essential part of algorithm development. The enormous dataset sizes and the complexity observed in many recent time-resolved 3D fluorescence microscopy imaging experiments, however, prohibit a comprehensive manual ground truth generation. Moreover, existing simulated benchmarks in this field are often too simple or too specialized to sufficiently validate the observed image analysis problems. We present a new semi-synthetic approach to generate realistic 3D+t benchmarks that combines challenging cellular movement dynamics of real embryos with simulated fluorescent nuclei and artificial image distortions including various parametrizable options like cell numbers, acquisition deficiencies or multiview simulations. We successfully applied the approach to simulate the development of a zebrafish embryo with thousands of cells over 14 hours of its early existence.
\end{abstract}
\section{Introduction}
\label{sec:Introduction}
The extensive use of benchmarking is essential for successful algorithmic development, particularly, to validate the successful operation of a new algorithm, to quantitatively compare existing and newly developed approaches, and to systematically optimize algorithmic performance. In recent years, various benchmarks for bioimage analysis have been presented for tasks such as seed detection \cite{Gelasca09, Ruusuvuori08}, segmentation \cite{Ruusuvuori08, Coelho09} or tracking \cite{Sbalzarini05, Rapoport11}. A general problem with manually created benchmark datasets, however, is caused by the inter- and intra-expert variability, which means that ambiguous image content may be rated differently by different investigators or even by the same investigator during multiple labeling iterations. An increasingly popular solution to tackle these problems and to additionally avoid time-consuming and tedious manual annotations is the use of simulated benchmark datasets. It has been shown that biological phenomena such as fluorescently labeled cell populations can be realistically simulated if enough knowledge of the investigated probes was available \cite{Lehmussola07, Svoboda12, Rapoport11, Maska14}. The charm of simulated data is the availability of a reliable ground truth and literally unrestricted possibilities to adjust parameters like noise levels, sampling rates or light attenuation, which can hardly be achieved by imaging dynamically changing organisms and thus prohibits robustness analyses as in \cite{Khan15}. Nevertheless, existing simulated benchmarks are often much simpler than the real application scenarios and mostly focus solely on a single processing step. Challenges such as multiview acquisition and fusion \cite{Preibisch10, Tomer12, Krzic12}, large file sizes \cite{Keller08SC, Stegmaier14} and highly dynamic scenes with possibly thousands of objects \cite{Amat14, Kobitski15}, that are frequently observed in state-of-the-art experiments in embryomics using confocal or light-sheet microscopy, are not considered sufficiently yet. To evaluate the performance of an entire image analysis pipeline comprised of seed detection, segmentation, multiview fusion and tracking with a single benchmark, we present a new method that combines simulated fluorescent objects, realistic object movement based on real embryos and the ability to generate challenging large-scale microscopy data in a single framework including various acquisition deficiencies. In the remainder of this paper, we introduce the general concept that was used to generate new benchmark datasets and a proof-of-principle simulation that mimics the early development of a zebrafish embryo.

\section{A New Semi-Synthetic Benchmark}
The new benchmark required a realistic simulation of fluorescence properties of labeled nuclei and a customizable number of cells, nucleus size, division cycle duration and experimental duration. The simulation of nuclei should include realistic cell movement, cell divisions, neighborhood related movement dynamics as well as spatial restrictions. Moreover, the generated simulation images should be artificially flawed by acquisition deficiencies such as an approximated point spread function (PSF), slice-dependent illumination variations, simulated multiview generation including light attenuation along the virtual axial direction as well as detector- and discretization-related deficiencies like dark current, photon shot noise and signal amplification noise. To achieve these requirements, we use object locations, displacement vectors and density information of real embryos and complemented the remaining components with synthetic data to a comprehensive benchmark generation framework offering the desired flexibility (Fig.\,1).
\begin{figure}[htb]
\centerline{\includegraphics[width=\columnwidth]{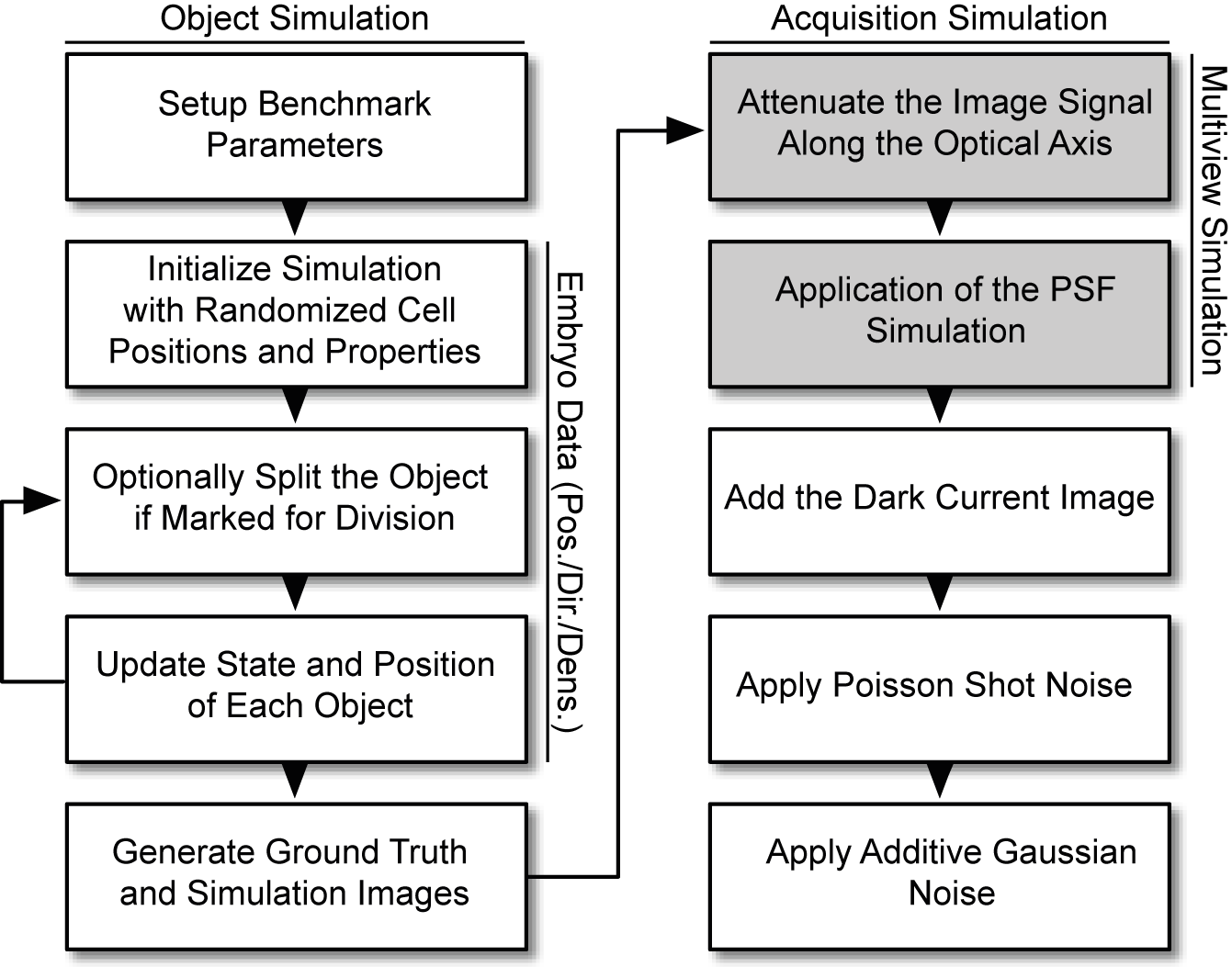}}
\caption[Pipeline schematic of the benchmark simulation]{Performed steps for a semi-synthetic 3D+t benchmark dataset using positions, directions and density variations of real embryos. The left column represents the object simulation steps and returns raw images that contain dynamic objects and associated ground truth data. The acquisition simulation (right column) distorts images by an artificial signal attenuation, a point spread function simulation (PSF), a dark current image simulation, Poisson distributed photon shot noise and additive Gaussian noise. Steps shaded in gray can optionally be adjusted to simulate multiview experiments.}
\label{fig:Figure1}
\end{figure}

\subsection{Benchmark Initialization}
The first step of the benchmark generation is to specify the number of objects that should be simulated. We use a percentage value $p\in [0,1]$ that specifies the relative amount of cells of the original organism that should be used for the simulation. The simulated objects are uniformly distributed in the object space using randomly selected spatial locations of real objects of the underlying embryo, \ie, the size of the simulated volume is determined by the extents of the underlying microscopy images that are used for the simulation. Alternatively, it is also possible to specify a fixed number of initial cells for the first frame instead of a relative number of cells, \eg, to guarantee comparable object counts even for different embryo datasets. In addition to the positioning of the initial objects, each object obtains a unique ID $i$, a radius $r_i \in [r_{\text{min}}, r_{\text{max}}]$, a division cycle length $l_i \in [l_{\text{min}}, l_{\text{max}}]$, a division cycle state $s_i \in [1,l_i]$ and an object video ID $o_i \in [1,N_{\text{ov}}]$, with $N_{\text{ov}}$ being the number of single-cell object videos. The subsequent simulation of the cellular movement is determined by these randomized parameters and the valid parameter ranges can be estimated by manually investigating a few representative cells of the original images.

\subsection{Cell Division Events}
One of the most challenging events that takes place during early embryonic development are cell divisions. There are various possibilities to add mitotic events to the simulation. The most straightforward approach is to simply specify minimum and maximum division cycle duration based on biological prior knowledge of the simulated specimen and to randomly assign a value between those two boundaries to each simulated cell. In each frame, the division cycle state is incremented and as soon as the maximum division cycle duration is reached, an object division is performed. However, this approach does not incorporate spatial information, \ie, all cells are dividing in a similar manner that does not necessarily correspond to real embryonic development. We found that the results get more realistic if the division events are directly coupled to the real number of objects. In each frame, the number of required cell divisions to reach the target number of cells is set to:
\begin{gather}
	N^{\text{div}}_k = \max (0, p \cdot N^{\text{embryo}}_k - N^{\text{sim}}_k).
	\label{eq:numDivisions}
\end{gather}
In Eq.\,(1), $p$ is the percentage of real cells used for the simulation and $N^{\text{embryo}}_k$ and $N^{\text{sim}}_k$ are the number of cells in the real and the simulated embryo at time point $k$. To identify which of the $N^{\text{sim}}_k$ cells should be divided, either the $N^{\text{div}}_k$ cells with the largest division cycle state are split or the divisions are performed density-based, by splitting the $N^{\text{div}}_k$ cells with $s_i \geq l_{\text{min}}$ with the largest relative density difference:
\begin{gather}
	\rho^{\text{diff}}_{ik} = \frac{\rho^\text{embryo}_{ik}}{N^{\text{embryo}}_k}  - \frac{\rho^\text{sim}_{ik}}{N^{\text{sim}}_k}.
	\label{eq:DensityDifference}
\end{gather}
The densities $\rho^\text{embryo}_{ik}$ and $\rho^\text{sim}_{ik}$ are the number of neighboring cells of object $i$ at time point $k$ within a fixed radius $r_\rho$ around each cell calculated either on the real data or on the simulated data. Although, the framework in principle allows using each of the described cell division approaches, the method based on the relative density difference yielded the most realistic results and was used for all presented results.

\subsection{Adding Simulated Object Dynamics}
The next step after the initialization and the selection of a cell division approach is the dynamic simulation. This step essentially comprises updating each object's spatial location as well as the simulated division cycle state. If an object's division cycle ended during the performed update step ($s_i \geq l_i$) or if it was selected for division due to a large relative density difference, an object division is performed. Each of the two new objects is again randomly initialized and positioned relative to its ancestor, with the division axis being set to the major axis of the mother cell.

To obtain a dynamically changing scene, the position of each object is updated at every simulation step by considering a set of simulated influences that are acting on it. The interactions are comprised of displacement vectors $\Delta\mathbf{x}^{\text{dir}}$, $\Delta\mathbf{x}^{\text{rep}}$ and $\Delta\mathbf{x}^{\text{nna}}$ that originate from real object movements of the underlying embryo, repulsive behavior between nearby objects and an attraction that pulls simulated objects towards the embryo, respectively. The displacement vector of the directed cell movement that is defined by the movement direction of real cells that reside in the vicinity of the simulated object. This is the most important component to obtain realistically moving objects and it is defined as:
\begin{gather}
	\Delta\mathbf{x}^{\text{dir}}(\mathbf{x}) = \frac{1}{K} \cdot \sum_{j \in \mathcal{N}^K_{\text{knn}}(\mathbf{x})}{\mathbf{d}_j},
\label{eq:chap4:Benchmark:CellMovement}
\end{gather}
where $K$ is the number of neighbors to use, $\mathcal{N}^K_{\text{knn}}(\mathbf{x})$ are the indices of the $K$ nearest neighbors of $\mathbf{x}$ and $\mathbf{d}_j$ is the movement direction of neighbor $j$. A repulsive component acting between two objects if their distance becomes smaller than the sum of their radii avoids intersections and is defined as:
\begin{align}
\Delta\mathbf{x}^{\text{rep}}&(\mathbf{d}) = \nonumber \\
 &\begin{cases} 
		-\left( c \cdot \frac{\Vert \mathbf{d} \Vert}{R_N}+1 \right) \cdot \frac{\mathbf{d}}{\Vert \mathbf{d} \Vert}, & 0 \leq \Vert \mathbf{d} \Vert \leq R_N \\
		- \left( 1 - \frac{\Vert \mathbf{d} \Vert}{R_M} \right)^2 \cdot \frac{\mathbf{d}}{\Vert \mathbf{d} \Vert}, & R_N < \Vert \mathbf{d} \Vert \leq R_M \\
				\mathbf{0}, & \text{else},
	 \end{cases}
	\label{eq:chap4:Benchmark:RepulsiveForce}
\end{align}
where
\begin{gather}
		c = \left( 1- R_N / R_M \right)^2-1.
\end{gather}
In Eq.\,(4), $\mathbf{d}=\mathbf{x}_j-\mathbf{x}_i$ is the centroid difference vector of two interacting objects $i,j$, $R_N$ is the radius of the cell nucleus \cite{Macklin12}. As the simulation was only performed on cell nuclei, the parameters were set in relation to the nucleus radii $r_i, r_j$ of two interacting objects $i, j$ to $R_N=r_i+r_j$ for the nucleus radius parameter and $R_M=2 \cdot R_N$ for the membrane radii. The repulsive displacement can in some cases push nuclei apart from the locations of the real embryo. To compensate this behavior, we additionally introduce a displacement vector component that slightly pulls each of the simulated objects towards its nearest neighbor $\mathcal{N}^1_{\text{knn}}(\mathbf{x})$:
\begin{gather}
		\Delta\mathbf{x}^{\text{nna}}(\mathbf{x}) = \mathcal{N}^1_{\text{knn}}(\mathbf{x}) - \mathbf{x}.
\end{gather}
The influence of the directed movement, the repulsive interaction and the nearest neighbor attraction can be controlled using the weights $w_{\text{dir}}$, $w_{\text{rep}}$ and $w_{\text{nna}}$, respectively. The total displacement vector of a single object at a given time point can be summarized to:
\begin{align}
	\Delta\mathbf{x}_i^{\text{tot}} = &w_{\text{dir}} \cdot \Delta\mathbf{x}^{\text{dir}}(\mathbf{x}_i) + \nonumber \\
	&w_{\text{rep}} \cdot \sum_{j\in \mathcal{N}^{r_{\text{max}}}_{\text{range}}(\mathbf{x})}{\Delta\mathbf{x}^{\text{rep}}(\mathbf{x}_j-\mathbf{x}_i)} + \nonumber \\
	\min ( & w_{\text{nna}}, \frac{\Vert \Delta\mathbf{x}^{\text{dir}}(\mathbf{x}_i) \Vert}{\Vert \Delta\mathbf{x}^{\text{nna}}(\mathbf{x}_i) \Vert} ) \cdot \Delta\mathbf{x}^{\text{nna}}(\mathbf{x}_i).
	\label{eq:chap4:Benchmark:TotalForce}
\end{align}
The nearest neighbor attraction weight $w_{\text{nna}}$ in Eq.\,(7) is clamped by the magnitude of the cell movement vector $\Delta\mathbf{x}^{\text{dir}}$ to avoid large jumps in cases where the nearest neighbor of a simulated object is erroneously missing in one or more frames of the underlying embryo data. Generally, it is possible to add further displacement components to the simulation, \eg, to add a specific attractor. However, for the benchmark only movements that originated from the real cells (Brownian-like and directed movements) and density variations caused by cell divisions were considered. 

\subsection{Simulation of Fluorescence Microscopy Images}
To generate the actual benchmark images and the corresponding label images from the simulated object locations, the images were initialized as entirely black images. Small single-cell 3D video sequences (Fig.\,2C) were extracted from a simulated time-lapse dataset comprised of eight dividing cells over two division cycles (data provided by D.~Svoboda \cite{Svoboda12}) and served as an object video database for the generation of the artificial benchmark images ($N_{\text{ov}}=56$). By iterating over all simulated time points and all simulated objects, both the benchmark images and the ground truth images were successively filled with simulated fluorescent nuclei and the label masks, respectively. The specified object radii and the division cycle lengths of the simulated objects were used to scale the single-cell videos appropriately. To simulate the acquisition process of a fluorescence microscope, all generated images were filtered in several steps to obtain a realistic benchmark dataset (Fig.\,1, right column). First, the intensities of the simulated images were attenuated along the virtual optical axis by multiplying the intensities of each slice by a linearly decreasing factor that is set to $1$ at the slice closest to the virtual detection objective and to $0$ at the slice farthest from the detection objective. Subsequently, the entire image was convolved with a point spread function (PSF) published in \cite{Preibisch14} that was measured by imaging fluorescent beads in a light-sheet microscope. To optionally simulate a multiview acquisition experiment with a single rotation of $180^\circ$, the multiplier used for signal attenuation was inverted and a point spread function that was analogously rotated by $180^\circ$ was used to convolve the images.
An empirically determined positive offset determined from fluorescence microscopy images was added to all intensity values, in order to simulate the dark current signal of the detector. To simulate photon shot noise, an independent Poisson process was applied to each voxel with the respective image intensities being its average \cite{Preibisch14}. Finally, a zero-mean additive Gaussian noise with a standard deviation of $\sigma_{\text{agn}}$ was used to model the readout noise caused by signal amplification.
The steps for modeling the acquisition deficiencies of a benchmark image ${\mathbf{I}^{\text{raw}}}$ are:
\begin{gather}
	\mathbf{I}^{\text{final}} = P_{\lambda}(\mathbf{I}^{\text{raw}} \ast \mathbf{I}^\text{psf} + \mathbf{I}^{\text{dark}}) + \mathcal{N}(0, \sigma_{\text{agn}}),
\end{gather}
where $\mathbf{I}^\text{psf}$ is the point spread function, $\mathbf{I}^{\text{dark}}$ is the dark current image of the detector, $P_\lambda$ applies a Poisson-based shot noise and finally $\mathcal{N}(0, \sigma_{\text{agn}})$ is a normally distributed random variable with zero mean and standard deviation $\sigma_{\text{agn}}$ \cite{Svoboda12}. The entire acquisition simulation was implemented in XPIWIT \cite{Bartschat15}.

\section{Simulating Early Zebrafish Development}
For the generation of an exemplary benchmark dataset, we used the spatio-temporal data of an early wild-type zebrafish embryo \cite{Stegmaier14, Kobitski15}. The displacement vector weights were set to $w_{\text{dir}}=1.0$, $w_{\text{rep}}=1.0$ and $w_{\text{nna}}=0.1$, and $K=10$ neighbors were used to estimate the object movements. The radius ranges were set to $r_{\text{min}}=7~\mu m$ and $r_{\text{max}}=10~\mu m$, respectively, with $N_{\text{ov}}=56$ different single-cell videos. We used a density-based cell division model where cell divisions were directly coupled to real amount of cells ($p \in \lbrace 0.25, 0.5, 0.75 \rbrace$, \ie, 25\%, 50\%, 75\% of the number of cells of the real embryo) and locations of cell divisions were determined via the maximum density difference (Eq.\,(2), $r_\rho=40~\mu m$, $l_{\text{min}}=28$). All parameters were empirically determined to obtain movement behaviors that nicely resembled the actual cellular dynamics observed during real embryonic development. It should be noted, though, that the presented model does not necessarily represent an accurate physical simulation of the interacting objects.

We successfully generated multiple time series of simulated embryos with varying numbers of cells that perfectly resemble the movement behavior and cell distributions of a real embryo (Fig.\,2A,B). On the basis of these simulated object locations, simulated fluorescent nuclei (Fig.\,2C, \cite{Svoboda12}) were used to generate time-resolved artificial benchmark images (Fig.\,2D). The generated data comprises ground truth label images, raw images including acquisition deficiencies and an object property database for each of the frames, and offers numerous possibilities to validate and analyze the robustness of images analysis and tracking operators.
\begin{figure}[htb!]
\centerline{\includegraphics[width=\columnwidth]{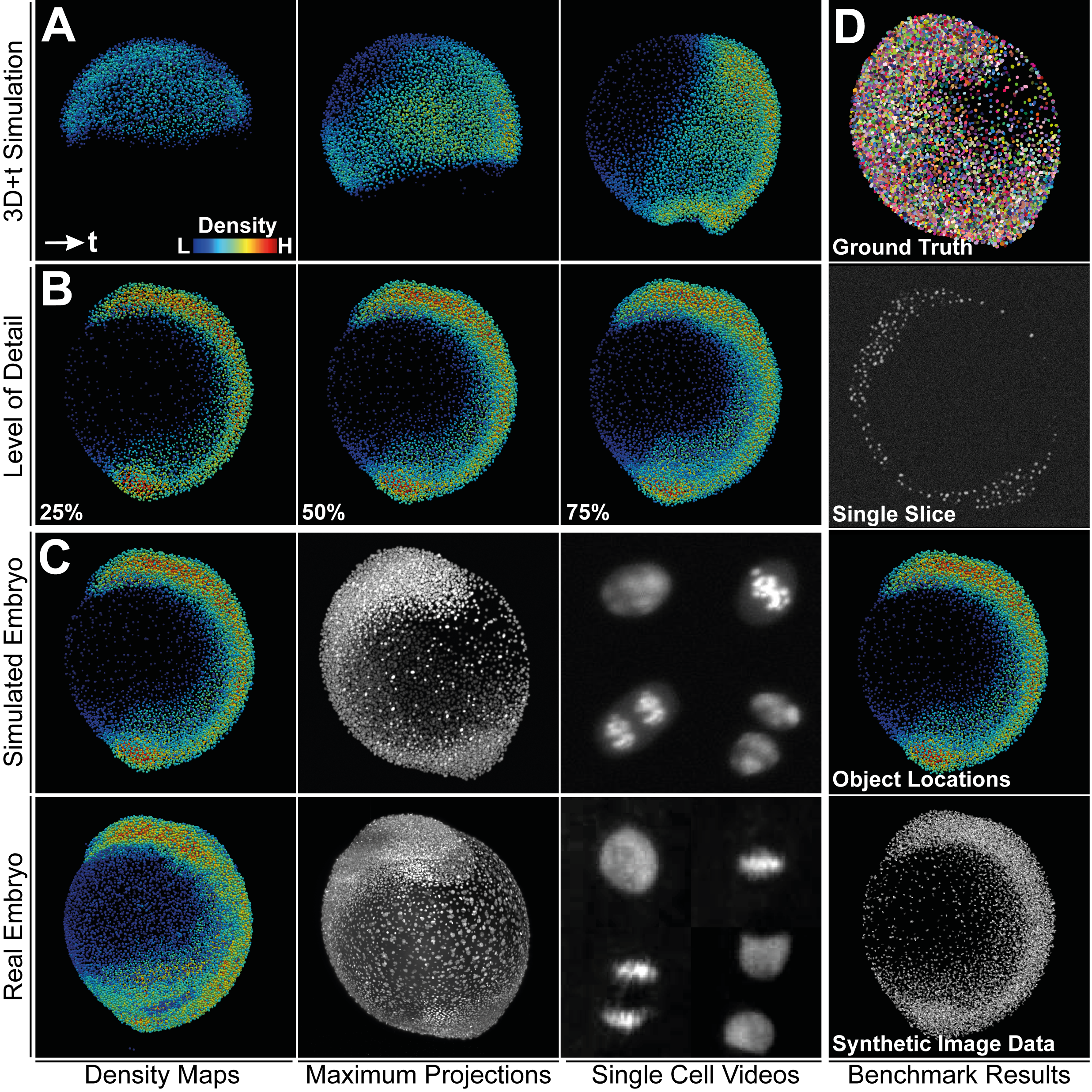}}
\caption[Maximum intensity projections of simulated benchmark images]{(A) Selected frames of a simulated embryo with 50\% of the cells of a real embryo (color encodes density). (B) A simulated embryo with different amounts of cells (25\%, 50\%, 75\%, color encodes density). (C) Comparison of simulated (top row) vs. real (bottom row) data using density maps, whole embryo maximum intensity projections and single cell videos of an extracted division cycle of one nucleus \cite{Svoboda12}. (D) The output of the simulation comprises label images, raw images, simulated microscopy images and meta information like object locations for each frame.}
\label{fig:Figure2}
\end{figure}

\section{Conclusions}
In this contribution we present a novel approach on how to generate realistic benchmark images that mimic the spatio-temporal cell dynamics of developing embryos. We successfully simulated the early development of a zebrafish embryo at various levels of detail that nicely imitated the movement behavior observed in a real embryo. Depending on the desired number of cells, the simulation currently takes a few hours and artificial images can be obtained in a matter of minutes for a single frame. The framework is currently implemented in MATLAB and we host source code, sample data and videos on \url{https://bitbucket.org/jstegmaier/}. We plan to extend the benchmark framework by an easy-to-use graphical user interface and to systematically speed-up the benchmark generation in order to easily produce benchmark datasets for numerous application fields and additional model organisms. Currently, the fluorescent objects used for the simulated images are based on artificial cells. A straightforward extension of the presented framework would be to replace the simulated video object library with manually annotated snippets of real microscopy images. If simulations should get even more realistic, cell distances could be learned from real data and tissue-dependent light scattering as well as a more realistic light attenuation model could be added.

%\section{Acknowledgements}
%We are grateful for funding by the Helmholtz Association in the program BioInterfaces (BS, RM), the Karlsruhe Institute of Technology (JA, PS) and the Deutsche Forschungsgemeinschaft (JS).

%% the bibliography
\bibliographystyle{IEEEbib}
%\bibliographystyle{diss}
%\bibliography{Bibliography}

\end{document}